%% file: ijcai24.tex
\documentclass{article}
\usepackage{ijcai24}

\usepackage{times}
\usepackage{soul}
\usepackage{url}
\usepackage[hidelinks]{hyperref}
\usepackage[utf8]{inputenc}
\usepackage[small]{caption}
\usepackage{graphicx}
\usepackage{amsmath}
\usepackage{amsthm}
\usepackage{booktabs}
\usepackage{algorithm}
\usepackage{algorithmic}
\usepackage[switch]{lineno}
\usepackage{amsfonts}
\usepackage{amssymb}

\title{Stochastic Adversarial Networks for Multi-Domain Text Classification}

\author{
Xu Wang$^1$
\And
Yuan Wu$^{1,2}$\footnote{Corresponding author.}\\
\affiliations
$^1$School of Artificial Intelligence, Jilin University\\
$^2$Key Laboratory of Symbolic Computation and Knowledge Engineering, Jilin University\\
\emails
xwang22@mails.jlu.edu.cn,
yuanwu@jlu.edu.cn
}

\begin{document}

\maketitle

\begin{abstract}

Adversarial training has been instrumental in advancing multi-domain text classification (MDTC). Traditionally, MDTC methods employ a shared-private paradigm, with a shared feature extractor for domain-invariant knowledge and individual private feature extractors for domain-specific knowledge. Despite achieving state-of-the-art results, these methods grapple with the escalating model parameters due to the continuous addition of new domains. To address this challenge, we introduce the Stochastic Adversarial Network (SAN), which innovatively models the parameters of the domain-specific feature extractor as a multivariate Gaussian distribution, as opposed to a traditional weight vector. This design allows for the generation of numerous domain-specific feature extractors without a substantial increase in model parameters, maintaining the model's size on par with that of a single domain-specific extractor. Furthermore, our approach integrates domain label smoothing and robust pseudo-label regularization to fortify the stability of adversarial training and to refine feature discriminability, respectively. The performance of our SAN, evaluated on two leading MDTC benchmarks, demonstrates its competitive edge against the current state-of-the-art methodologies. The code is available at \url{https://github.com/wangxu0820/SAN}.
\end{abstract}

\section{Introduction}
\input{sec-introduction}

\section{Related Work}
\input{sec-related-work}

\section{Method}
\input{sec-method}

\section{Experiments}
\input{sec-experiments}

\section{Conclusion}
In this study, we introduce a Stochastic Adversarial Network (SAN) specifically devised for MDTC tasks. Our approach distinctively models the weights of domain-specific feature extractors through a multivariate Gaussian distribution $\mathcal{N} (\mu,\Sigma)$. This design allows for network weights to be sampled directly from the distribution when employing domain-specific feature extractors. A notable advantage of this methodology is its capacity to minimize the model's parameter count, preventing parameter escalation with the addition of new domains. Additionally, we incorporate domain label smoothing and robust pseudo-label regularization to ensure stable adversarial training and to enhance feature discrimination. Our experimental evaluation, conducted on two MDTC benchmarks, validates the SAN model's capability to improve system performance and demonstrates its robust generalization to unfamiliar domains.

\bibliographystyle{named}
\bibliography{ijcai24}

\appendix

\section{Center of samples on sphere}
This section computes the class center of spherical samples used for robust pseudo-label regularization. 
Before computing the class centers on sphere, we begin by normalizing the concatenations of the shared features and domain-specific features. Let $f(\textbf{x})=[F_s(\textbf{x}),F_d(\textbf{x})]$, where $[\cdot,\cdot]$ represents the concatenation of two vectors. We then normalize features with $f'=r \frac{f(\textbf{x})}{f(\textbf{x})}$ to obtain features in the spherical space $\mathbb{S}_r^{n-1}=\{f'\in\mathbb{R}^n:||f'||=r\}$. Let $f'_1,f'_2,\cdots,f'_m$ be samples on the sphere $\mathbb{S}_r^{n-1}$, the center $\mathcal{C}$ of the samples on the sphere corresponds to the point closest to all samples, i.e., the solution of the following optimization problem:

\begin{equation}\label{eq156}
    \min \limits_{f'\in\mathbb{S}_r^{n-1}}\frac{1}{m}\sum_{i=1}^m dist(f',f'_i)
\end{equation}
    
Where $dist(u,v)=1-\frac{u^{T}v}{||u||||v||}$ is the cosine distance. Since $||f'||=r, \forall f'\in\mathbb{S}_r^{n-1}$, Eq. \ref{eq156} can be rewritten as:

\begin{equation}
    \max \limits_{f'} f'^T(\sum_{i=1}^m f'_i) \hspace{1cm} s.t. ||f'||=r.
\end{equation}

With the method of Lagrange multipliers, the center can be obtained by:

\begin{equation}
    \mathcal{C}=\frac{r}{||\tilde{f'}||} \tilde{f'}
\end{equation}

Where $\tilde{f'}=\sum_{i=1}^mf'_i$.

\section{How stochastic feature extractor works}

\begin{figure}[h]
    \centering
    \includegraphics[width=1\columnwidth]{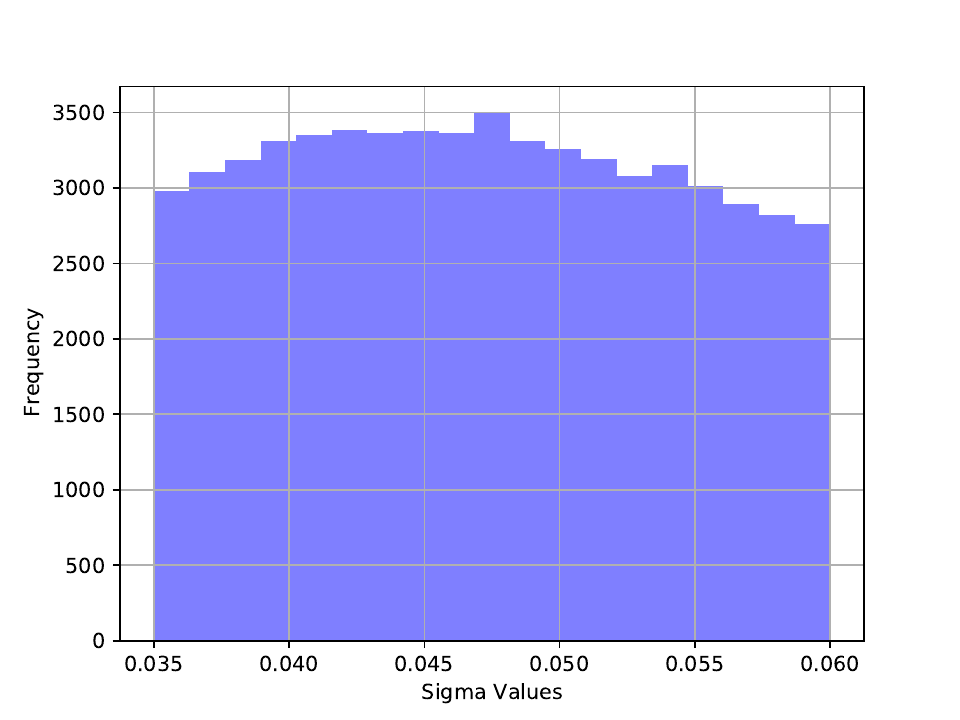}
    \caption{The distribution of the flattened $\Sigma$ values for initialization on the Amazon Review dataset.}
    \label{fig-epoch-1}
\end{figure}

\begin{figure}[h]
    \centering
    \includegraphics[width=1\columnwidth]{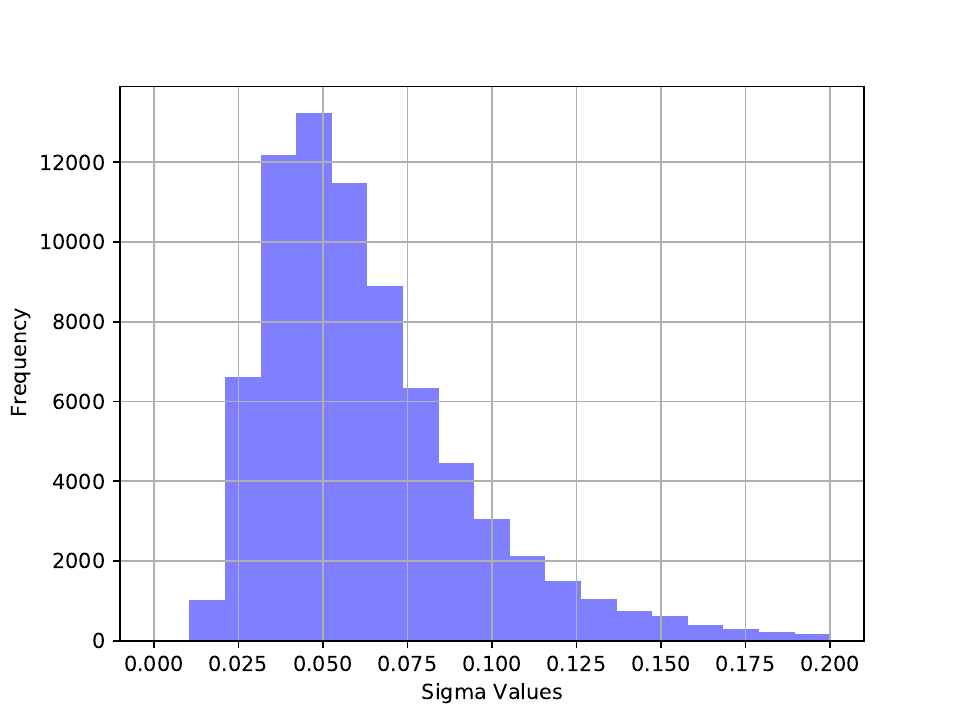}
    \caption{The distribution of the flattened $\Sigma$ values after convergence of SAN on the Amazon Review dataset.}
    \label{fig-epoch-30}
\end{figure}

The variance of the distribution, denoted as \(\Sigma\), plays a crucial role in the operational dynamics of the stochastic feature extractor. As illustrated in~\figurename~\ref{fig-epoch-1}, the initial values of \(\Sigma\) are distributed uniformly, manifesting no discernible structure. However, post-training, these values start to display more defined patterns in~\figurename~\ref{fig-epoch-30} . It's particularly noteworthy that the distribution of the domain-specific feature extractors tends to show larger variances for different domains. As the SAN model progresses towards convergence, these pronounced variances act as a mechanism to guarantee the extraction and preservation of distinct, domain-unique features across the various domains, reinforcing the model's ability to handle domain-specific nuances effectively.

\section{Experiments}

\subsection{Dataset}

The experiments are conducted on two benchmark datasets: the Amazon review dataset~\cite{blitzer2007biographies} and the FDU-MTL dataset~\cite{liu2017adversarial}. The statistical specifics of these datasets, which are instrumental for our analysis, are concisely presented in~\tablename~\ref{tb-data-amazon} for the Amazon review dataset and~\tablename~\ref{tb-data-fdu} for the FDU-MTL dataset.

\input{Appendix/tables/tb-data-amazon}
\input{Appendix/tables/tb-data-fdu}

\subsection{Validity verification of stochastic feature extractor}
\input{Appendix/tables/tb-verify}

We meticulously designed a suite of experiments leveraging the Amazon review dataset to empirically validate the capability of the SAN model to adeptly learn domain-specific features from multiple domains. 
We rigorously evaluate three variants of the model:
(1) pSAN: Represents the foundational SAN model without any modifications.
(2) pSAN w/ \(F_{d}^{zero}\): A variant in which domain-specific features are intentionally set to zero, thereby exclusively relying on shared features for classification.
(3) pSAN w/ \(F_{d}^{shuf}\): This variant introduces a permutation in the domain-specific features across different domains, thereby simulating the scenario where domain-specific features from one domain are input into another.
The empirical results clearly indicate that pSAN w/ \(F_{d}^{shuf}\) not only lags behind the baseline pSAN but also performs inferiorly compared to pSAN w/ \(F_{d}^{zero}\). 
This phenomenon distinctly suggests that arbitrarily shuffled domain-specific features do not contribute constructively to the classification task. Instead, they act as confounding variables, detracting from the model's overall efficacy. This observation underscores the SAN model's intrinsic ability to extract and leverage salient domain-specific features effectively.

\subsection{Parameter sensitivity analysis}
\begin{figure}[h]
    \centering
    \includegraphics[width=1\columnwidth]{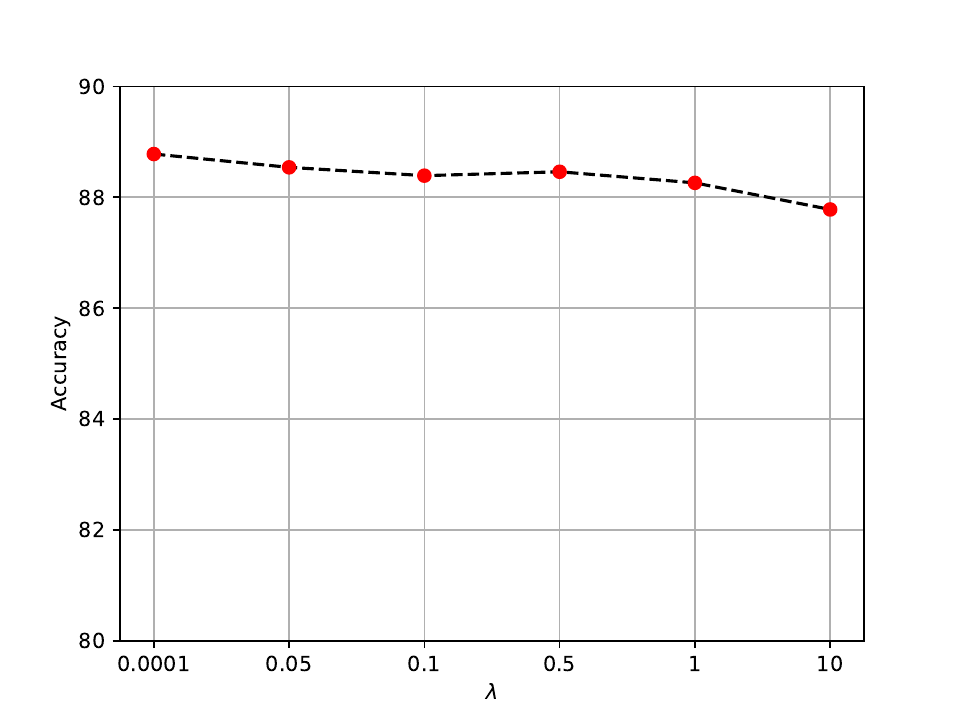}
    \caption{Parameter sensitivity analysis about $\lambda$ on Amazon review dataset.}
    \label{fig-sa-lambda}
\end{figure}

\begin{figure}[h]
    \centering
    \includegraphics[width=1\columnwidth]{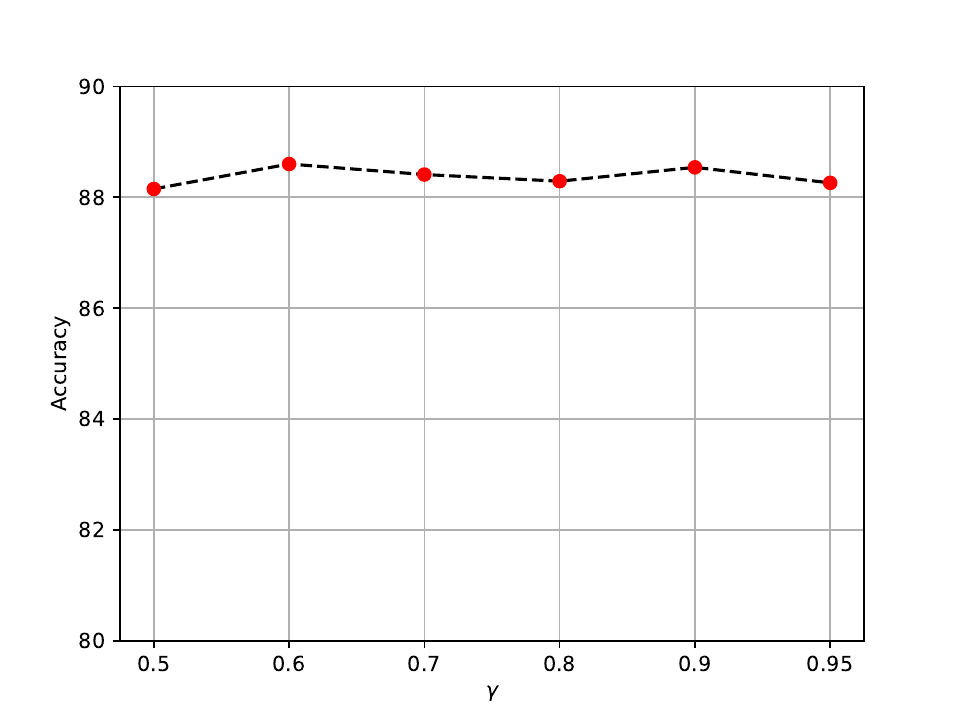}
    \caption{Parameter sensitivity analysis about $\gamma$ on Amazon review dataset.}
    \label{fig-sa-gamma}
\end{figure}

\begin{figure}[h]
    \centering
    \includegraphics[width=1\columnwidth]{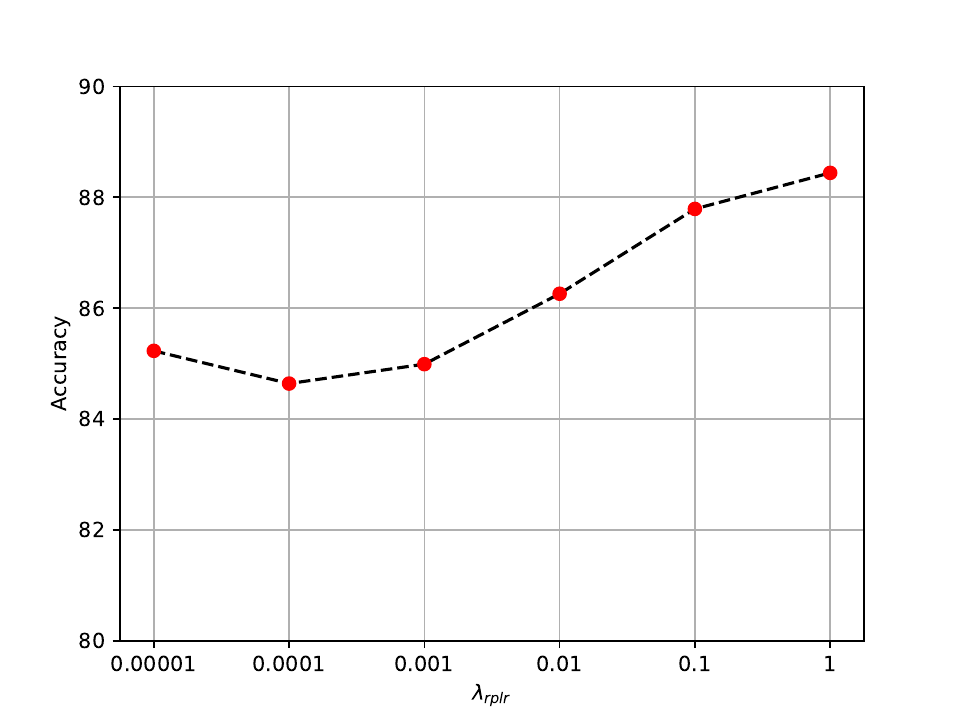}
    \caption{Parameter sensitivity analysis about $\lambda_{rplr}$ on Amazon review dataset.}
    \label{fig-sa-lambda-rplr}
\end{figure}

In this section, we examine the sensitivity of our SAN method to the values of hyperparameters $\lambda$, $\gamma$ and $\lambda_{rplr}$. The $\lambda$ and $\lambda_{rplr}$ are evaluated in the range $\left \{ 0.0001,0.05,0.1,0.5,1,10 \right \} $ and $\left \{ 0.00001,0.0001,0.001,0.01,0.1,1 \right \} $, respectively. The valid range of values for $\gamma$ is $\left ( 0,1 \right ] $, therefore we assess its impacts in the range of $\left \{ 0.5,0.6,0.7,0.8,0.9,0.95 \right \}$. 
This comprehensive parameter sensitivity analysis is conducted utilizing both the Amazon review dataset and the FDU-MTL dataset. We visually present the outcomes of this investigation in~\figurename~\ref{fig-sa-lambda},~\ref{fig-sa-gamma} and~\ref{fig-sa-lambda-rplr}, respectively, focusing on the metric of average classification accuracy.

\subsection{Ablation study}

To discern the individual impact of each constituent in our SAN methodology on the overall performance, we embark on an ablation study, leveraging both the Amazon review dataset and the FDU-MTL dataset. The outcomes of this analysis are systematically illustrated in~\tablename~\ref{tb-ablation-amazon} for the Amazon dataset and~\tablename~\ref{tb-ablation-fdu} for the FDU-MTL dataset.
Specifically, we examine three variants: 
(1) SAN w/o dls, a variant without the enhancement of domain label smoothing; 
(2) SAN w/o rplr, a variant without the enhancement of robust pseudo-label regularization; 
(3) plain SAN, a variant utilizing a stochastic feature extractor instead of the domain-specific feature extractors of shared-private scheme.
The findings from each variant underscore the integral role of the individual components, with all three variants demonstrating diminished performance compared to the comprehensive model. Notably, the full SAN model, integrating all components, consistently delivers the most superior performance, conclusively affirming the collective contribution of both domain label smoothing and robust pseudo-label regularization to the enhancement of our model's performance.

\input{Appendix/tables/tb-ablation-amazon}
\input{Appendix/tables/tb-ablation-fdu}

\subsection{Convergence analysis}

\begin{figure}[h]
    \centering
    \includegraphics[width=1\columnwidth]{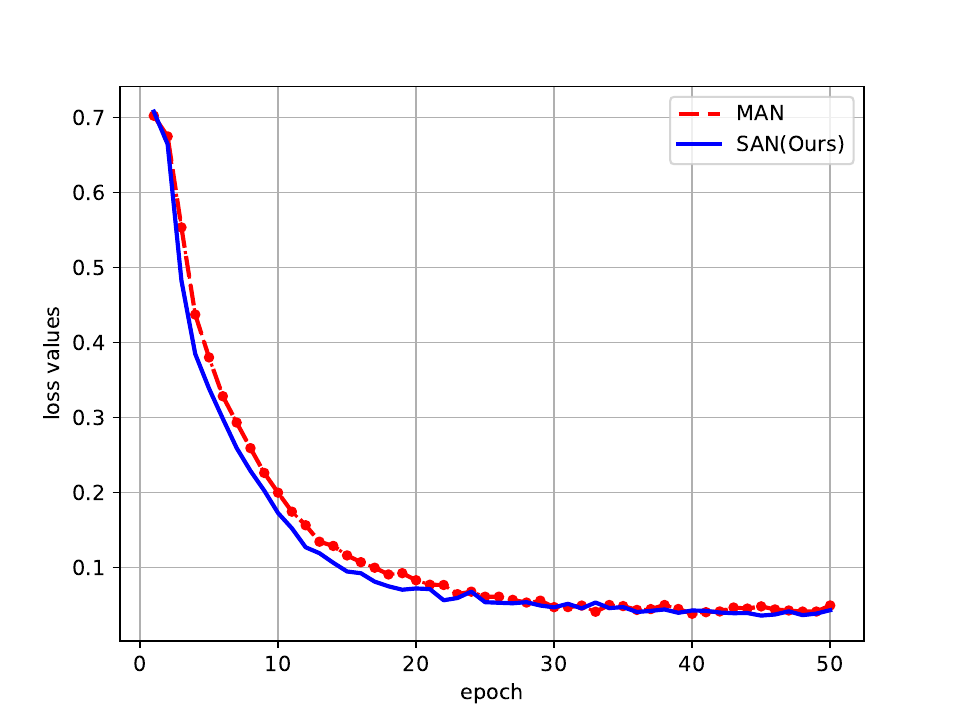}
    \caption{Convergence analysis between MAN and SAN(ours).}
    \label{fig-loss}
\end{figure}

We conducted a comparative analysis of the convergence speed between our novel SAN model and conventional MDTC methods that utilize the shared-private framework, exemplified by the Multinomial Adversarial Networks (MAN) as referenced in~\cite{chen2018multinomial}. The comparative results, as delineated in~\figurename~\ref{fig-loss}, underscore the superior convergence rate of our SAN approach relative to MAN. This enhanced convergence efficiency can be attributed to the innovative implementation of the stochastic feature extractor in our SAN model, which not only streamlines the model's learning process but also expedites its rate of convergence.

\input{Appendix/tables/tb-limit-comparison}
\input{Appendix/tables/tb-limit-amzon}
\input{Appendix/tables/tb-limit-fdu}

\section{Limitations}

While our SAN model exhibits commendable performance on the Amazon review dataset, its efficacy on the FDU-MTL dataset does not match that of the leading MDTC methods. We present a comparative analysis in~\tablename~\ref{tb-limit-comparision}, comparing our SAN model with some of the most recent MDTC methodologies, including the Conditional Adversarial Network (CAN)~\cite{wu2021conditional}, the Mixup Regularized Adversarial Network (MRAN)~\cite{wu2021mixup}, the Co-Regularized Adversarial Network (CRAL)~\cite{wu2022co}, the Robust Contrastive Alignment (RCA)~\cite{li2022robust}, and the Maximum Batch Frobenius Norm (MBF)~\cite{wu2022maximum}.

A significant constraint identified in our approach pertains to the less-than-optimal accuracy of the pseudo-labels employed during the robust pseudo-label regularization process. Within the framework of the SAN model, a pseudo-labeled data point $\textbf{x}^u_j$ is deemed valid if its weight $w(\textbf{x}^u_j)$ surpasses 0.5.
The accuracy of these valid pseudo-labels on the unlabeled data from the Amazon review dataset is detailed in~\tablename~\ref{tb-limit-amazon}. Furthermore,~\tablename~\ref{tb-limit-fdu} delineates the accuracy of valid pseudo-labels on both the validation and test sets of the FDU-MTL dataset. A notable observation from this data is the considerable discrepancy in pseudo-label accuracy across various domains within the FDU-MTL dataset, with the 'MR' domain notably attaining only 82.87\% accuracy.
This disparity in pseudo-label quality can significantly undermine the overall performance of the system. Consequently, we posit that a pivotal area for enhancing the efficacy of our SAN model lies in improving the precision of pseudo-labels assigned to the unlabeled data, a move that is anticipated to substantially elevate the model's performance metrics.

\end{document}

%% file: sec-introduction.tex
Text classification has become a prominent area of focus within the field of Natural Language Processing (NLP)~\cite{khurana2023natural}. The preceding decade has seen remarkable progress in deep learning, which has significantly propelled the capabilities of text classification~\cite{kowsari2019text,zhou2024survey}. There's a broad consensus that textual content is intrinsically domain-specific~\cite{wu2022maximum}. This means a single word might evoke different sentiments in different contexts, leading to situations where a model, though effective in its training domain, may not yield optimal results in a new, untrained domain. The endeavor to amass extensive, labeled datasets for each specific domain often proves to be prohibitively expensive and challenging. Hence, it is imperative to explore and develop methods capable of harnessing knowledge from related domains to enhance the accuracy of classification in the target domain.

Multi-domain text classification (MDTC) is proposed to address the challenges highlighted above~\cite{li2008multi,hu2024regularized}. 
Initial MDTC methods relied on domain-specific training and employed ensemble learning to produce final results~\cite{li2012multi,wu2015collaborative}. 
However, the latest MDTC approaches, which utilize adversarial training~\cite{creswell2018generative,ganin2016domain} and a shared-private scheme~\cite{bousmalis2016domain}, deliver state-of-the-art performance. 
Adversarial training is utilized to align distinct domains, thereby facilitating the extraction of domain-invariant features. The shared-private scheme splits the latent space into two parts: a shared space that captures common features across domains, and a private space dedicated to capturing unique features specific to each domain. The domain-invariant features are expected to be both discriminative and transferable across domains, whereas the domain-specific features are intended to augment the distinctiveness and discriminative power of the feature set~\cite{bousmalis2016domain}.
However, these methods encounter a notable challenge. The shared-private framework requires the construction of domain-specific feature extractors for each domain, often involving complex neural networks. As new domains are introduced, the addition of numerous domain-specific extractors not only increases the model's complexity but also impedes training convergence.

To address the challenges outlined, we introduce a novel framework termed Stochastic Adversarial Network (SAN), which employs a stochastic feature extractor as a replacement for multiple domain-specific feature extractors. This innovative extractor amalgamates an unlimited number of domain-specific extractors into prevailing MDTC methodologies without altering the model's parameter count. In SAN, the conventional practice of utilizing 
specific weight points is replaced with a weight distribution, signifying the domain-specific feature extractors. 
Specifically, we employ a Gaussian distribution to model these extractors, with the mean symbolizing the central weight of the domain-specific feature extractor and the variance denoting the discrepancy across distinct domains. 
Throughout the training phase, the domain-specific feature extractor is periodically sampled from the prevailing distribution estimate, concurrently optimizing the Gaussian distribution. As a result, SAN is proficient in extracting domain-specific features from numerous domains utilizing a singular extractor. 
This approach circumvents the substantial escalation in the model's parameter count that typically accompanies an increase in the number of domains, thereby ensuring the model size remains stable. 
To further refine model performance, we integrate domain label smoothing and robust pseudo-label regularization within the SAN. This integration promotes stability during adversarial training and enhances the discriminative capability of features. Empirical evaluations on two established MDTC benchmarks substantiate the efficacy of our SAN model, achieving competitive performance compared to state-of-the-art methods.

Our contributions are summarized as follows:
\begin{itemize}
\item We propose the Stochastic Adversarial Network (SAN) for MDTC, introducing an innovative stochastic feature extractor mechanism. This mechanism facilitates the extraction of domain-specific features across various domains through a singular stochastic extractor, substantially reducing the model's parameter count. To the best of our knowledge, this study represents the first exploration of this matter in MDTC.
\item We incorporate domain label smoothing and robust pseudo-label regularization techniques to stabilize the adversarial training and enhance the discriminability of the acquired features, respectively.
\item Experimental results on two benchmark datasets highlight the effectiveness of the SAN approach relative to state-of-the-art methods. Additionally, a comparative analysis of the number of parameters and running time between SAN and conventional MDTC methods showcases the superior efficiency of our proposed approach.
\end{itemize}

%% file: sec-related-work.tex
\textbf{Adversarial Training (AT)}, initially conceptualized within the Generative Adversarial Network (GAN) framework for image generation, involves a dual mechanism: a generator creating images and a discriminator differentiating between synthesized and authentic images~\cite{creswell2018generative}. 
Domain-Adversarial Neural Networks (DANN) extend AT to domain adaptation, training a feature extractor to counter a domain discriminator~\cite{ganin2016domain}. The discriminator strives to identify source and target features, while the feature extractor seeks to produce domain-invariant features undetectable by the discriminator. 
Conditional Adversarial Neural Networks (CDAN) further advance this approach by applying multilinear conditioning to synchronize conditional distributions and incorporating entropy conditioning to aid transfer learning~\cite{long2018conditional}. 
However, AT is not without challenges; it's prone to oscillatory gradients during training, leading to issues such as instability, delayed convergence, and mode collapse~\cite{arjovsky2017towards,mescheder2018training}. 
To mitigate these issues, Wasserstein GAN leverages the earth mover distance for a more refined domain divergence measure~\cite{arjovsky2017wasserstein}. 
Moreover, Environment Label Smoothing (ELS) is employed to prompt the domain discriminator to generate soft probabilities, thereby enhancing AT's stability~\cite{zhang2023free}.

\begin{figure*}[t]
    \centering
    \includegraphics[width=1.8\columnwidth]{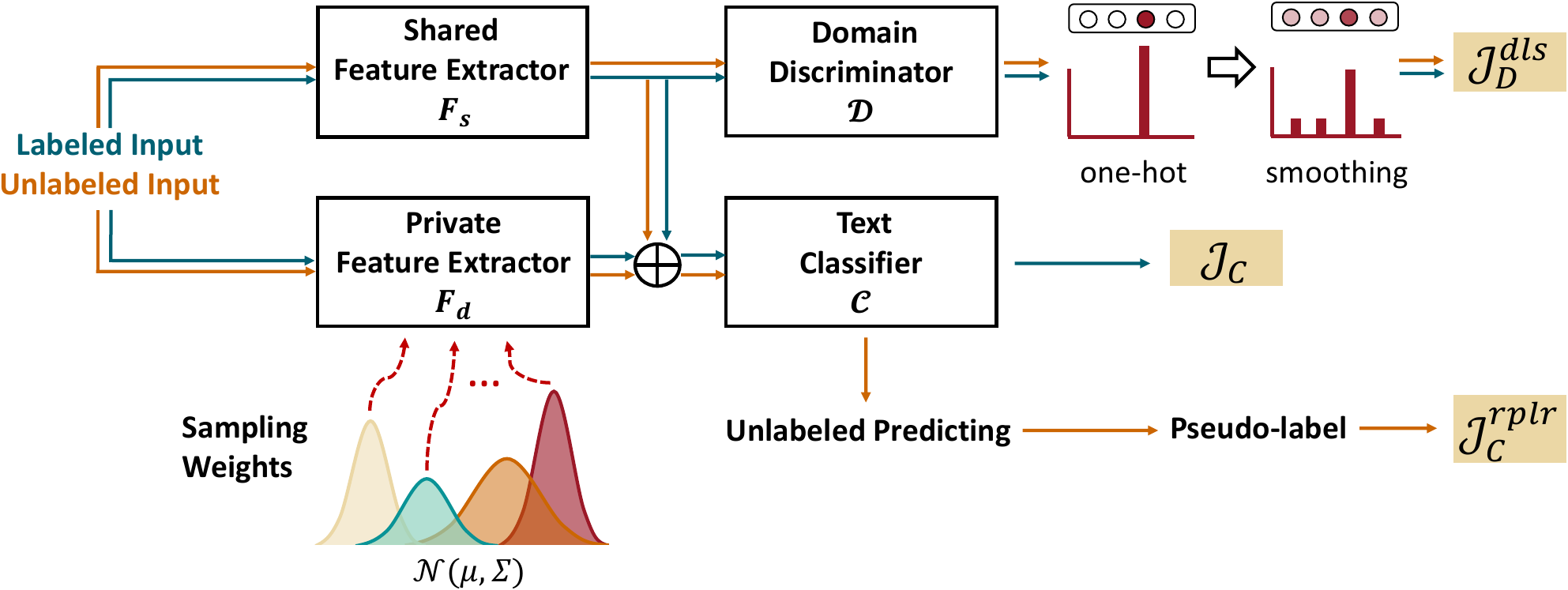}
    \caption{The architecture of the SAN method.
	}
    \label{fig-structure}
\end{figure*}

\textbf{Stochastic Neural Network (SNN).} The weight parameters of a neural network are typically treated as point estimates, limiting their ability to capture uncertainty and often resulting in overconfident predictions~\cite{blundell2015weight}. To address this limitation, SNNs are proposed, which consider weight parameters as random variables sampled from specific distributions. For example, Bayesian Neural Networks (BNNs)~\cite{hernandez2015probabilistic,wang2020survey} are widely used to represent intermediate outputs and final predictions as stochastic variables, providing richer representations. The Auto-Encoding Variational Bayes (AEVB)~\cite{kingma2013auto} employs a Gaussian distribution to model latent variables in image inputs, serving as a form of data augmentation. Uncertainty-aware multi-modal BNNs~\cite{subedar2019uncertainty} combine deterministic and variational layers for activity recognition, while DistributionNet~\cite{yu2019robust} models feature uncertainty in person re-identification using distributions. In unsupervised domain adaptation, the Stochastic Classifier~\cite{lu2020stochastic} leverages a Gaussian distribution to model classifier parameters.

\textbf{Multi-domain text classifications (MDTC).}  MDTC aims to enhance overall classification accuracy by harnessing available resources from multiple domains~\cite{li2008multi}. Early MDTC methods employ transfer learning techniques to drive progress. The structural correspondence learning (SCL)~\cite{blitzer2006domain} method computes relationships between different pivot features to learn correspondences among them. The collaborative multi-domain sentiment classification (CMSC)~\cite{wu2015collaborative} method trains two types of classifiers: a shared classifier for all domains and a set of domain-specific classifiers for each domain, combining their outputs for final results. Recent MDTC approaches commonly adopt the adversarial training and shared-private paradigm, leading to significant advancements. The domain separation network (DSN)~\cite{bousmalis2016domain} first introduces the shared-private paradigm for adversarial domain adaptation and empirically demonstrates that domain-unique features can enhance the discriminability of domain-invariant features. The adversarial multi-task learning (ASP-MTL) method~\cite{liu2017adversarial} applies adversarial training and the shared-private paradigm to MDTC. The multinomial adversarial networks (MAN)~\cite{chen2018multinomial} utilize the least square loss and negative log-likelihood loss to train the domain discriminator. The mixup regularized adversarial networks (MRANs)~\cite{wu2021mixup} propose domain and category mixup regularizers for MDTC. The maximum batch Frobenius norm (MBF)~\cite{wu2022maximum} method improves feature discriminability by maximizing the Frobenius norm of the intermediate feature matrix.

Our proposed SAN method contrasts with traditional MDTC techniques that deploy distinct domain-specific feature extractors for each domain. Instead, SAN adopts a parameter sampling strategy from a Gaussian distribution to instantiate domain-specific feature extractors. This innovative approach allows SAN to obtain domain-specific insights through a stochastic feature extractor, resulting in a significant reduction in the number of model parameters needed.

%% file: sec-method.tex
The MDTC task can be formulated as follows: given $M$ domains $\left\{D_{i}\right\}^{M}_{i = 1}$, each domain contains a small amount of labeled data $\mathbb{L}_{i}  =\left \{ \textbf{x}_{j},y_{j}  \right \} _{j=1}^{l_{i}}$ and a large amount of unlabeled data $\mathbb{U}_{i}  =\left \{ \textbf{x}_{j}  \right \} _{j=1}^{u_{i}}$. The primary objective of MDTC is to leverage these resources to enhance the average classification accuracy across all domains.

\subsection{Adversarial Multi-Domain Text Classification}

Adversarial training has garnered recognition for effectively mitigating domain discrepancies, and its application in MDTC is increasingly prevalent~\cite{chen2018multinomial,wu2020dual,wu2022maximum}.
Conventional adversarial MDTC frameworks typically encompass four key components: 
(1) a shared feature extractor $F_s$,
(2) an array of domain-specific feature extractors $\{F_d^i\}_{i=1}^M$,
(3) a classifier $C$, and (4) a domain discriminator $D$. 
The primary role of $F_s$ is to distill features that are invariant to the domain, while $\{F_d^i\}_{i=1}^M$ are specialized to capture distinctive features that are uniquely advantageous within their specific domains. The classifier $C$ is responsible for sentiment prediction, and $D$ discerns the domain of the input, thereby aiding in domain adaptation. 
Feature extractors can be instantiated using a variety of neural network architectures, including Convolutional Neural Networks (CNNs)~\cite{zhang2015character}, Multi-Layer Perceptrons (MLPs)~\cite{chen2018multinomial}, and Transformers~\cite{vaswani2017attention}. These architectures are proficient in generating fixed-length feature representations from input data. 
In this configuration, $D$ is fed the shared feature vector, while $C$ utilizes a concatenation of the shared feature vector and the domain-specific feature vector for its predictions.

In traditional MDTC paradigms, the primary goals involve 
(1) minimizing the classification loss on labeled data to ensure accurate predictions, and (2) concurrently optimizing the adversarial loss on both labeled and unlabeled data to facilitate effective domain adaptation. These objectives are typically formulated as follows:

\begin{align}
    \min\limits_{F_s,\{F_d^i\}_{i=1}^M,C}\max\limits_{D}\mathcal{J}_{C}(F_s,\{F_d^i\}_{i=1}^M,C)+\lambda\mathcal{J}_{D}(F_s,D)
\end{align}
\begin{equation}
\label{eq1}
    \mathcal{J}_{C}(F_s,\{F_d^i\}_{i=1}^M,C)= \sum_{i=1}^{M}\mathbb{E}_{(\textbf{x},y)\sim \mathbb{L}_{i}} [\mathcal{L}(C[F_{s}(\textbf{x}),F_{d}^i(\textbf{x})],y)]
\end{equation}
\begin{equation}
    \mathcal{J}_{D}(F_s,D)= \sum_{i=1}^{M}\mathbb{E}_{\textbf{x}\sim \mathbb{L}_{i}\cup \mathbb{U}_{i} } [\mathcal{L}(D(F_{s}(\textbf{x})),d)]
\end{equation}

Where $\mathcal{L}(\cdot,\cdot)$ is the loss function, $[\cdot,\cdot]$ represents the concatenation of two vectors, and $d$ is the ground-truth domain label of the corresponding instance $\textbf{x}$. 

\subsection{Stochastic adversarial network}
Given the complexity of neural network architectures employed by feature extractors to distill valuable insights from input data, and the necessity for MDTC models to train individual domain-specific feature extractors for each domain, this conventional approach often results in a significant increase in the model's parameter count and a deceleration in convergence speed. To address these challenges, we introduce the Stochastic Adversarial Network (SAN) tailored for MDTC. SAN innovates by integrating a stochastic feature extractor, effectively supplanting the need for multiple domain-specific feature extractors without sacrificing model performance. The architecture of our proposed SAN method is illustrated in~\figurename~\ref{fig-structure}.
The cornerstone of our methodology is to model a distribution representing domain-specific feature extractors. In this model, the domain-specific feature extractors, which are pivotal for learning unique features within each domain, are not fixed entities but random samples drawn from this predefined distribution. This innovative approach affords the flexibility to access an infinite array of domain-specific feature extractors, as we can sample any desired number of extractors based on our requirements. Crucially, it also decouples the number of domain from the model parameter count, ensuring that the model's size remains stable.

In our approach, we adopt a multivariate Gaussian distribution, denoted as $\mathcal{N} (\mu,\Sigma)$, where $\mu$ represents the mean vector and $\Sigma$ represents to the diagonal covariance matrix. This distribution serves as the basis for generating the parameters of the domain-specific feature extractors for each domain, which are randomly sampled from $\mathcal{N} (\mu,\Sigma)$. Following the sampling, the incurred loss is back-propagated to refine the learnable parameters $\mu$ and $\Sigma$. 
It is important to note that the stochastic nature of the random sampling process disrupts the traditional flow of end-to-end training. To circumvent this impediment, we employ the reparameterization trick, as delineated in~\cite{kingma2013auto}, which enables efficient training of the model through backpropagation.
More specifically, we represent the last fully connected layer of the domain-specific feature extractor as $\phi_d$. This layer is formulated as $\phi_d = \mu + \sigma \odot \epsilon$, where $\epsilon$ is a random sample drawn from a standard Gaussian distribution, $\odot$ denotes element-wise multiplication, and $\sigma$ represents the diagonal elements of $\Sigma$.

By adopting the stochastic feature extractor, we can update Eq.\ref{eq1} and adversarial training as:

\begin{align}
\label{eq-san}
    \mathcal{J}_{C}(F_s,F_d,C)= \sum_{i=1}^{M}\mathbb{E}_{(\textbf{x},y)\sim \mathbb{L}_{i}} [\mathcal{L}(C[F_{s}(\textbf{x}),F_{d}(\textbf{x})],y)]
\end{align}


\subsection{Enhancement via domain label smoothing}

\begin{figure}
  \begin{center}
    \includegraphics[width=0.45\textwidth]{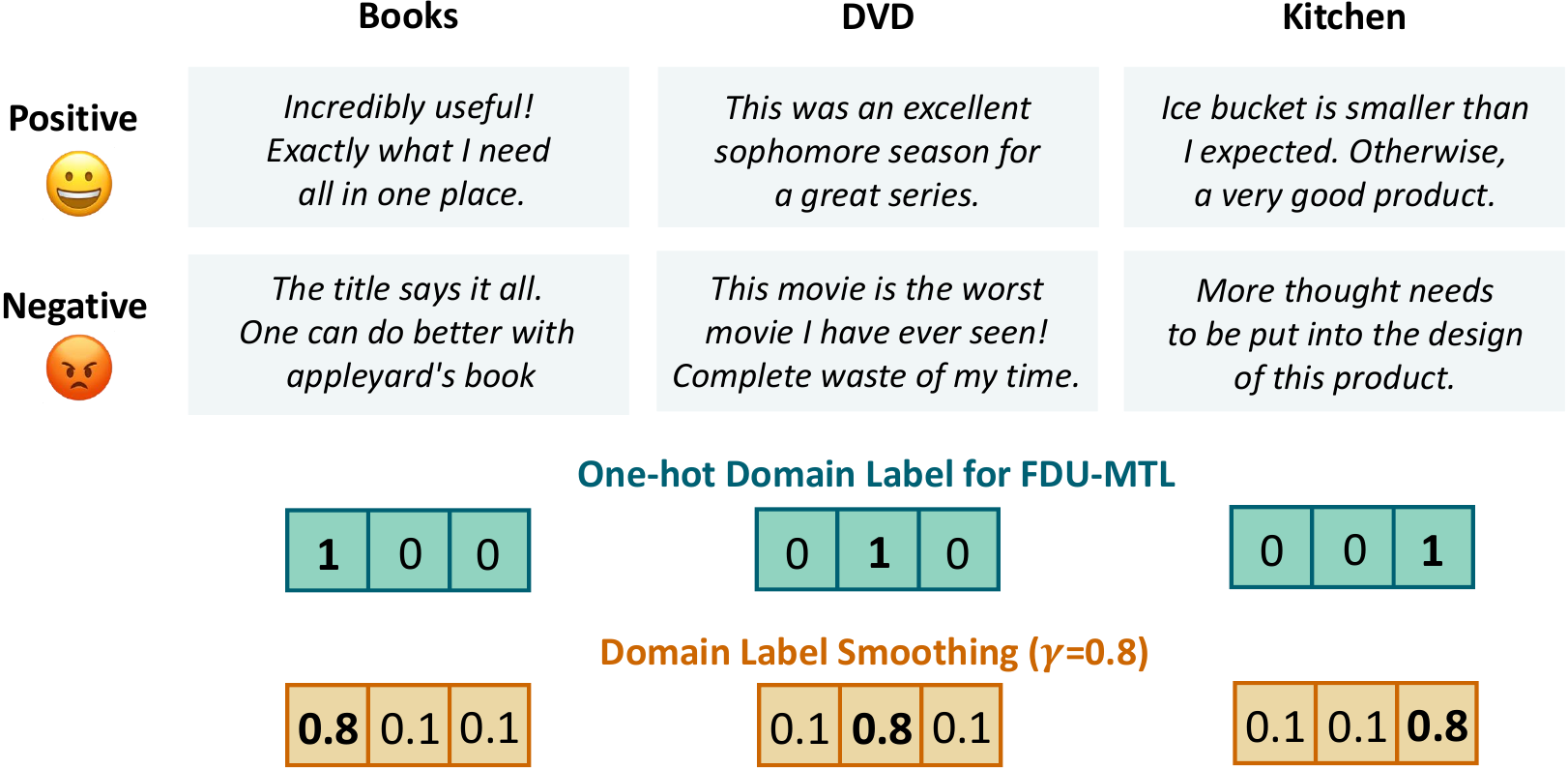}
  \end{center}
  \caption{An example of DLS with three domains on the FDU-MTL dataset.}
  \label{fig-dls}
\end{figure}

While Adversarial Training (AT) has been empirically validated for its effectiveness in minimizing domain divergence and capturing domain-invariant features~\cite{ganin2016domain,chen2018multinomial}, it is widely recognized that AT models can be challenging to train and may struggle to converge~\cite{roth2017stabilizing,jenni2019stabilizing,arjovsky2017towards}. This challenge primarily stems from the utilization of one-hot domain labels, which tends to lead to highly over-confident output probabilities. This over-confidence in the domain discriminator can generate substantial oscillatory gradients, adversely affecting the stability of the training process~\cite{arjovsky2017towards,mescheder2018training}.
To address the challenge of overconfidence in domain predictions, our model incorporates a technique referred to as Domain Label Smoothing (DLS), as depicted in~\figurename~\ref{fig-dls}. This approach is designed to temper the domain discriminator's predictions, shifting from absolute and potentially overconfident classifications to the estimation of softer, more nuanced probabilities~\cite{zhang2023free}.
The DLS formulation is as follows:

\begin{align}
\label{eq-dls}
    \mathcal{J}_{D}^{els}(F_s,D) = & \sum_{i=1}^{M}\mathbb{E}_{\textbf{x}\sim \mathbb{L}_{i}\cup \mathbb{U}_{i}} [\gamma \log(\mathcal{D}_i(F_{s}(\textbf{x}))) \notag \\
    & +\frac{1-\gamma}{M-1} \sum_{j=1,j\neq i}^{M} \log(\mathcal{D}_j(F_{s}(\textbf{x})))],
\end{align}

Where $\mathcal{D}_i$ gives the $i$-th dimension of the domain discriminator's output vector and $\gamma$ ($\gamma\in(0,1)$) is a hyperparameter. 
DLS is both theoretically and empirically proven to bolster the model's robustness against noisy domain labels and accelerate convergence. It promotes stable training and superior generalization performance, all without necessitating extra parameters or additional optimization steps.

With Eq.\ref{eq-san} and Eq.\ref{eq-dls}, the adversarial training can be updated as:
\begin{align}\label{eq1111}
    \min\limits_{F_s,F_d,C}\max\limits_{D}\mathcal{J}_C(F_s,F_d,C)+\lambda\mathcal{J}_D^{els}(F_s,D)
\end{align}

\subsection{Enhancement via robust pseudo-label regularization}

The abundance of unlabeled data in each domain presents an opportunity to utilize pseudo-labels to refine feature discriminability in MDTC task. However, not all unlabeled data contribute positively. To judiciously select the unlabeled data that can yield dependable pseudo-labels, we incorporate the Robust Pseudo Label Regularization (RPLR) technique into our proposed SAN framework~\cite{gu2020spherical}. 
RPLR operates by evaluating the reliability of pseudo-labels for unlabeled data, gauging this based on the feature distance to the corresponding class center within a spherical feature space. It identifies incorrectly labeled data as outliers and employs a Gaussian-uniform mixture model to characterize the conditional probability of a data point being an outlier or an inlier.
For an input instance $\textbf{x}^u_j$ its generated pseudo-label $\hat{y}^u_j$ is defined as: $\hat{y}^u_j=\arg \max \limits_{k} [C[F_{s}(\textbf{x}^u_j),F_{d}(\textbf{x}^u_j)]]_{k}$, where $[\cdot]_k$ denotes the $k$-th element.
To assess the accuracy of the generated pseudo-label, we introduce a binary random variable $z_j\in\{0,1\}$, indicating the correctness of the labeling with 1 for correct and 0 for incorrect. 
RPLR is then articulated as follows:

\begin{align}        
 \mathcal{J}_{C}^{rplr}&(F_s,F_d,C,\phi)=   \notag \\
& \sum_{i=1}^{M}\mathbb{E}_{\textbf{x}^u_j\sim\mathbb{U}_{i}}[w(\textbf{x}^u_j)\mathcal{L}(C[F_s(\textbf{x}^u_j),F_d(\textbf{x}^u_j)],\hat{y}^u_j)]
\end{align}

\begin{align}
w(\textbf{x}^u_j)=
\begin{cases}
  \beta_j& \text{ if } \beta_j>0.5\\
  0& \text{ otherwise } 
\end{cases}
\end{align}

Where $\beta_j$ represents the probability of correctly labeled data, i.e., $\beta=Pr(z_j=1|\textbf{x}^u_j,\hat{y}^u_j)$. In this manner, unlabeled data with a probability of correct labeling below 0.5 are discarded. The posterior probability of correct labeling, i.e., $Pr(z=1|\textbf{x}_j,\hat{y}_j),$ is modeled by the feature distance between the data and the class center to which it belongs, using a Gaussian-uniform mixture model based on pseudo-labels. Given a feature vector $f^u_j=[F_s(\textbf{x}^u_j),F_d(\textbf{x}^u_j)]$ of an unlabeled instance $\textbf{x}^u_j$, its distance to the corresponding class center $\mathcal{C}_{\hat{y}^u_j}$ for category $\hat{y}^u_j$ is calculated as:

\begin{equation}
d^u_j=\frac{f^{u}_j\cdot \mathcal{C}_{\hat{y}^u_{j} } }{\left \| f^{u}_j\right \|  \left \| \mathcal{C}_{\hat{y}^u_{j}} \right \| } 
\end{equation}

The class center $\mathcal{C}_{\hat{y}^u_{j}}$ is defined in a spherical space as presented in \cite{gu2020spherical}, the details of computing $\mathcal{C}_{\hat{y}^u_{j}}$ are available in the \textbf{Appendix}. The distribution of feature distance $d^u_j$ is modeled by the Gaussian-uniform mixture model, a statistical distribution considering outliers \cite{coretto2016robust,lathuiliere2018deepgum}.
\begin{equation}
\label{eq6}
p(d^u_j|\hat{y}^u_j)=\pi_{\hat{y}^u_{j}} \mathcal{N}^{+}(d^u_j|0, \sigma_{\hat{y}^u_{j}})+ (1-\pi_{\hat{y}^u_{j}})\mathcal{U}(0,\delta _{\hat{y}^u_{j}}) 
\end{equation}
 
Where $\mathcal{N}^{+}(d^u_j|0, \sigma)$ denotes a density function that is proportional to Gaussian distribution when $d^u_j\ge0$, otherwise the density is zero. $\mathcal{U}(0,\delta_{\hat{y}^u_j})$ is uniform distribution defined on $[0,\delta_{\hat{y}^u_j}]$. Specifically, the Gaussian component captures the underlying probability distribution of correctly labeled data, while the uniform component provides a robust representation of the distribution for incorrectly labeled data. With Eq.\ref{eq6}, the posterior probability of correct labeling for unlabeled data $\textbf{x}_j$ is defined:

\begin{equation}
\beta_{j}= \frac{\pi _{\hat{y}^u_{j}} \mathcal{N}^{+}(d^u_{j}|0, \sigma_{\hat{y}^u_{j}})}{p(d^u_{j}|\hat{y}^u_{j})} 
\end{equation}

The parameters of Gaussian-uniform mixture models are $\phi=\{\pi_k,\sigma_k,\delta_k\}_{k=1}^K$ where $K$ is the number of classes. The details of approximating these parameters will be given in Sec.~\ref{sec-training}.

In summary, the ultimate optimization objective is defined as:
\begin{align}
\label{eq7}
    \min \limits_{F_s,F_d,C} \max \limits_{D} \mathcal{J}_C(F_s,F_d,C)+\lambda\mathcal{J}_D^{els}(F_s,D) \notag \\
    +\lambda_{rplr}\mathcal{J}_C^{rplr}(F_s,F_d,C,\phi)
\end{align}

\subsection{Training procedure}
\label{sec-training}

In this section, we present how to optimize each component in the SAN model and estimate the parameters $\phi$ of Gaussian-uniform mixture models. To optimize the ultimate object in Eq.\ref{eq7}, we alternatively optimize the networks and estimate parameters $\phi$ by fixing other components following \cite{gu2020spherical}. We first initialize $F_s$, $F_d$, $C$, $D$ with Eq.\ref{eq1111} via training strategies as in \cite{chen2018multinomial}, then we take the following two steps to make the optimization. 

\textbf{(1) Estimating $\phi$ with fixed $F_s$, $F_d$, $C$, $D$}. Fixing the parameters of $F_s$, $F_d$, $C$, $D$, we generate the pseudo-label $\hat{y}^u_j$ and calculate the distance $d^u_j$ for all unlabeled data, then $\phi$ is estimated using EM algorithm as below. Let $\tilde{d}^u_j=(-1)^m_j d^u_j$, where $m_j$ is sampled from Bernoulli distribution $B(1,0.5)$, and $N^u$ denotes the number of unlabeled data, then $\phi$ can be estimated as follows:

\begin{align}
    \beta^{l+1}_j=\frac{\pi^{l}_{\hat{y}^u_j}\mathcal{N}(\tilde{d}^u_j|0,\sigma^l_{\hat{y}^u_l})}{\pi^l_{\hat{y}^u_j}\mathcal{N}(\tilde{d}^u_j|0,\sigma^l_{\hat{y}^u_j})+(1-\pi^l_{\hat{y}^u_j})\mathcal{U}(-\delta^l_{\hat{y}^u_j},\delta^l_{\hat{y}^u_j})} \nonumber
\end{align}

\begin{align}
    \pi^{l+1}_k=\frac{1}{\sum_{j=1}^{N^u}I_{\{\hat{y}^u_j=k\}}}\sum_{j=1}^{N^u}I_{\{\hat{y}^u_j=k\}}\beta_j^{l+1} \nonumber
\end{align}

\begin{align}
    \sigma_j^{l+1}=\frac{\sum_{j=1}^{N^u}I_{\{\hat{y}^u_j=k\}}\beta_j^{l+1}(\tilde{d}^u_j)^2}{\sum_{j=1}^{N^u}I_{\{\hat{y}^u_j=k\}}\beta_j^{l+1}}, \delta_k^{l+1}=\sqrt{3(q_2-q_1^2)}\nonumber
\end{align}

Where 
\begin{align}
    q_1=\frac{1}{\sum_{j=1}^{N^u}I_{\{\hat{y}^u_j=k\}}\beta_j^{l+1}}\sum_{j=1}^{N^u}\frac{1-\beta_j^{l+1}}{1-\pi_k^{l+1}}I_{\{\hat{y}^u_j=k\}}\tilde{d}^u_j \nonumber
\end{align}

\begin{align}
    q_2=\frac{1}{\sum_{j=1}^{N^u}I_{\{\hat{y}^u_j=k\}}\beta_j^{l+1}}\sum_{j=1}^{N^u}\frac{1-\beta_j^{l+1}}{1-\pi_k^{l+1}}I_{\{\hat{y}^u_j=k\}}(\tilde{d}^u_j)^2 \nonumber
\end{align}

We refer our readers to \cite{gu2020spherical} for the deduction details of the parameters $\phi$. 

\textbf{(2) Optimizing $F_s$, $F_d$, $C$, $D$ with fixed $\phi$}. Given current pseudo-labels and estimated $\phi$, we follow the standard MDTC training protocol \cite{chen2018multinomial} to train $F_s$, $F_d$, $C$, $D$ with Eq.\ref{eq7}.

%% file: sec-experiments.tex
\input{tables/tb-amazon}

\input{tables/tb-fdu}

This section will expand on the three aspects of experiments setup, experiments results, and efficiency analysis.
\subsection{Setup}

\textbf{Datasets.} We conducted experiments on two benchmark datasets for MDTC task: the Amazon review dataset \cite{blitzer2007biographies} and the FDU-MTL dataset \cite{liu2017adversarial}. The Amazon review dataset comprises four domains: books, DVDs, electronics, and kitchen. Each domain consists of 2000 labeled data instances, with 1000 positive and 1000 negative examples. The data has been pre-processed into a bag-of-features representation, which includes unigrams and bigrams, without preserving word order information. The FDU-MTL dataset reflects real-world scenarios, as it contains raw text data. It encompasses 14 product review domains, including books, electronics, DVDs, kitchen, apparel, camera, health, music, toys, video, baby, magazine, software, sport, as well as two movie review domains: IMDB and MR. Each domain includes a validation set of 200 samples and a test set of 400 samples. The number of samples in the training and unlabeled sets varies across domains, but generally consists of approximately 1400 and 2000 instances, respectively. 

\input{tables/tb-uda}

\textbf{Implementation details.} 
To ensure a fair comparison, we adopt almost identical network architectures as presented in~\cite{chen2018multinomial}. 
It's pertinent to highlight that the sole modification we introduce is the substitution of the last fully connected layer of the original domain-specific feature extractor with a stochastic layer. 
For the Amazon review dataset, we select the 5000 most frequent features and represent each review as a 5000-dimensional vector, where the feature values represent raw counts. Our feature extractors employ multi-layer perceptrons (MLPs) with an input size of 5000. Each feature extractor consists of two hidden layers with sizes of 1000 and 500, respectively. In the case of the FDU-MTL dataset, we employ a single-layer convolutional neural network (CNN) as the feature extractor. The CNN utilizes different kernel sizes (3, 4, 5) with a total of 200 kernels. The input to the CNN is a 100-dimensional embedding obtained by processing each word of the input sequence using word2vec~\cite{mikolov2013efficient}. For all experiments, we set the batch size to 8, the dropout rate for each component to 0.4, and the learning rate of the Adam optimizer~\cite{kingma2014adam} to 0.0001. The size of the shared features is set to 128, and the size of the domain-specific features is set to 64. Both the classifier and discriminator are MLPs with hidden layer sizes matching their respective inputs (128+64 for the classifier and 128 for the domain discriminator). Furthermore, we set the hyperparameters $\lambda$ to 0.0001, $\gamma$ to 0.9, and $\lambda_{rplr}$ to 1.

\textbf{Comparison methods.} 
In the MDTC tasks, we evaluate the our SAN method against several state-of-the-art methods: The multi-task convolutional neural network (MT-CNN)~\cite{collobert2008unified}, the muti-task deep neural network (MT-DNN)~\cite{liu2015representation}, the collaborative multi-domain sentiment classification method (CMSC) trained with the least square loss (CMSC-LS), the hinge loss (CMSC-SVM), the log loss (CMSC-Log)~\cite{wu2015collaborative}, the pre-trained BERT-base model fine-tuned on each domain (BERT) \cite{devlin2018bert}, the adversarial multi-task learning for text classification method (ASP-MTL) \cite{liu2017adversarial}, the multinomial adversarial network (MAN) trained with the least square loss (MAN-L2) and the negative log-likelihood loss (MAN-NLL) \cite{chen2018multinomial}, the dynamic attentive sentence encoding method (DA-MTL) \cite{zheng2018same}, the global and local shared representation-based dual-channel multi-task learning method (GLR-MTL) \cite{su2020multi}, the conditional adversarial network (CAN) \cite{wu2021conditional}, and the co-regularized adversarial learning method \cite{wu2022co}. 
For MS-UDA experiments, the baselines involve the marginalized denoising autoencoder (mSDA)~\cite{chen2012marginalized}, the domain adversarial neural network \cite{ganin2016domain}, the multi-source domain adaptation network (MDAN)~\cite{zhao2017multiple}, the MAN (MAN-L2 and MAN-NLL)~\cite{chen2018multinomial}, the CAN~\cite{wu2021conditional} and CRAL~\cite{wu2022co}.

\subsection{Result}

\textbf{Multi-Domain Text Classification.} The experimental results on the Amazon review dataset and FDU-MTL dataset are reported in~\tablename~\ref{tb-amazon} and~\tablename~\ref{tb-fdu}, respectively. We report the classification results of mean $\pm$ variance over five random runs. From Table~\ref{tb-amazon}, it can be noted that the SAN method obtains the best classification accuracy on 3 out of 4 domains, and yield state-of-the-art results for the average classification accuracy. For the experimental results on FDU-MTL, shown in Table~\ref{tb-fdu}, the proposed SAN method outperforms MT-CNN and MT-DNN consistently across all domains with notable large performance gains. When compared with the state-of-the-art MAN-L2, MAN-NLL, DA-MTL, and GLR-MTL, SAN achieves competitive results in terms of average classification accuracy. The experimental results on both benchmarks validate the efficacy of our proposed method.

\textbf{Multi-Source Unsupervised Domain Adaptation.} In real application scenarios, it is not uncommon for the target domain to lack annotated data. Evaluating MDTC models under such circumstances is of utmost significance. In the multi-source unsupervised domain adaptation (MS-UDA) setting, we have multiple source domains, each containing both labeled and unlabeled data, and a target domain with only unlabeled data. Our MS-UDA experiments are conducted on the Amazon review dataset, following the same protocol as outlined in \cite{chen2018multinomial}. Specifically, in each experiment, three out of four domains were treated as source domains, while the remaining domain was treated as the target domain. As shown in~\tablename~\ref{tb-uda}, the proposed SAN method outperforms other baselines on two out of four domains as well as the average accuracy. It reveals that our SAN method has a good capacity for transferring knowledge to unseen domains. Further experimental results, including parameter sensitivity analysis, ablation study and convergence speed analysis can be found in the \textbf{Appendix}.

\subsection{Efficiency analysis}

To demonstrate the efficiency of our SAN model, this section provides a comparative analysis concentrating on two critical metrics: the number of model parameters and the running time. We compare traditional MDTC methods, notably exemplified by MAN~\cite{chen2018multinomial} and based on the shared-private paradigm, with our SAN approach. 

\textbf{Model parameter comparison.} 
We quantify the parameter counts for the shared feature extractor $F_s$, domain-specific feature extractors $\{F_d^i\}_{i=1}^M$, classifier $C$, and domain discriminator $D$ within both our SAN model and the traditional MAN model. The results, presented in~\tablename~\ref{tb-para-comparision}, illuminate a significant distinction: while the domain-specific feature extractors substantially contribute to the overall parameter count in traditional MDTC models, our SAN markedly reduces the parameter load attributed to these extractors.

\input{tables/tb-para-comparision}

\textbf{Model runtime comparison.} 
We also compared the runtime of our SAN model and MAN on the Amazon review and FDU-MTL datasets, using the average training time per epoch as the indicator. The results, which are summarized in~\tablename~\ref{tb-runtime}, are as follows: it is easy to observe that SAN requires less time, saving nearly 10\% compared to MAN on the Amazon dataset and nearly 15\% compared to MAN on the FDU-MTL dataset. 
This further validates the effectiveness of our SAN approach. 

\input{tables/tb-runtime}

%% file: tables/tb-amazon.tex
\begin{table*}[t]
\centering
\resizebox{1.9\columnwidth}{!}{
\begin{tabular}{ l|  c c c c c c c c c}
\toprule
Domain & CMSC-LS & CMSC-SVM & CMSC-Log & MAN-L2 & MAN-NLL & CAN & CRAL &SAN(Ours)\\
\midrule
books &  82.10 & 82.26 & 81.81 & 82.46 & 82.98 & 83.76 &85.26 & $\mathbf{86.29\pm0.26}$ \\
DVD &  82.40 & 83.48 & 83.73 & 83.98 & 84.03 & 84.68 &85.83 & $\mathbf{86.43\pm0.38}$ \\
electronics & 86.12 & 86.76 & 86.67 & 87.22 & 87.06 & 88.34 &89.32 & $\mathbf{89.78\pm0.12}$ \\
kitchen  &  87.56 & 88.20 & 88.23 & 88.53 & 88.57 & 90.03 & $\mathbf{91.60}$ &91.31$\pm0.15$\\
\midrule
AVG  &  84.55 & 85.18 & 85.11 & 85.55 & 85.66 & 86.70 &88.00 & $\mathbf{88.45\pm0.08}$\\
\bottomrule
\end{tabular}}
\caption{MDTC results on the Amazon review dataset.}
\label{tb-amazon}
\end{table*}

%% file: tables/tb-fdu.tex
\begin{table*}[t]
\centering
\resizebox{1.9\columnwidth}{!}{
\begin{tabular}{ l| c c c c c c c c c c}
\toprule
Domain & MT-CNN & MT-DNN & ASP-MTL & BERT & MAN-L2 & MAN-NLL & DA-MTL & GLR-MTL & SAN(Ours)\\
\midrule
books & 84.5 & 82.2 & 84.0 & 87.0 & 87.6 & 86.8 & 88.5 & 88.3 &  $\mathbf{90.5\pm0.3}$ \\
electronics & 83.2 & 88.3 & 86.8 & 88.3 & 87.4 & 88.8 & 89.0 & $\mathbf{90.3}$ & 87.7$\pm$0.6 \\
dvd & 84.0 & 84.2 & 85.5 & 85.6 & 88.1 & 88.6 & 88.0 & 87.3 & $\mathbf{89.7\pm0.5}$ \\
kitchen & 83.2 & 80.7 & 86.2 & $\mathbf{91.0}$ & 89.8 & 89.9 & 89.0 & 89.8 & 90.4$\pm$0.9 \\
apparel & 83.7 & 85.0 & 87.0 & $\mathbf{90.0}$ & 87.6 & 87.6 & 88.8 & 88.2 & 87.4$\pm$0.7\\
camera & 86.0 & 86.2 & 89.2 & 90.0 & $\mathbf{91.4}$ & 90.7 & 91.8 & 89.5 & 91.1$\pm$0.6 \\
health & 87.2 & 85.7 & 88.2 & 88.3 & 89.8 & 89.4 & 90.3 & $\mathbf{90.5}$ & 90.3$\pm$0.3 \\
music & 83.7 & 84.7 & 82.5 & 86.8 & 85.9 & 85.5 & 85.0 & $\mathbf{87.5}$ & 85.9$\pm$0.8 \\
toys & 89.2 & 87.7 & 88.0 & 91.3 & 90.0 & $\mathbf{90.4}$ & 89.5 & 89.8 & 90.3$\pm$0.7 \\
video & 81.5 & 85.0 & 84.5 & 88.0 & 89.5 & 89.6 & 89.5 & $\mathbf{90.8}$ & 90.0$\pm$0.5 \\
baby & 87.7 & 88.0 & 88.2 & 91.5 & 90.0 & 90.2 & 90.5 & $\mathbf{92.3}$ & 90.7$\pm$0.8 \\
magazine & 87.7 & 89.5 & 92.2 & $\mathbf{92.8}$ & 92.5 & 92.9 & 92.0 & 92.3 & 92.3$\pm$0.1\\ 
software & 86.5 & 85.7 & 87.2 & 89.3 & 90.4 & 90.9 & 90.8 & $\mathbf{91.8}$ & 89.5$\pm$0.4 \\
sports & 84.0 & 83.2 & 85.7 & 90.8 & 89.0 & 89.0 & $\mathbf{89.8}$ & 87.8 & $\mathbf{90.0\pm0.2}$ \\
IMDb & 86.2 & 83.2 & 85.5 & 85.8 & 86.6 & 87.0 & 89.8 & 87.5 &89.3$\pm0.7$\\
MR & 74.5 & 75.5 & $\mathbf{76.7}$ & 74.0 & 76.1 & 76.7 & 75.5 & 72.7 & 76.5$\pm0.9$\\
\midrule
AVG & 84.5 & 84.3 & 86.1 & 88.1 & 88.2 & 88.4 & 88.2 & 88.5 & $\mathbf{88.8\pm0.1}$ \\
\bottomrule
\end{tabular} 
}
\caption{MDTC results on the FDU-MTL dataset.}
\label{tb-fdu}
\end{table*}

%% file: tables/tb-uda.tex
\begin{table*}[h]
\centering
\resizebox{1.80\columnwidth}{!}{
\begin{tabular}{ l| c c c c c c c c c}
\toprule
Domain & mSDA & DANN & MDAN(H) & MDAN(S) & MAN-L2 & MAN-NLL & CAN & CRAL & SAN(Ours) \\
\midrule
books & 76.98 & 77.89 & 78.45 & 78.63 & 78.45 & 77.78 & 78.91 & $\mathbf{82.49}$ & 81.48\\
DVD & 78.61 & 78.86 & 77.97 & 80.65 & 81.57 & 82.74 & 83.37 &84.30 & $\mathbf{85.53}$\\
electronics & 81.98 & 84.91 & 84.83 & 85.34 & 83.37 & 83.75 & 84.76 &86.82 &$\mathbf{87.12}$\\
kitchen & 84.26 & 86.39 & 85.80 & 86.26 & 85.57 & 86.41 & 86.75 & $\mathbf{89.08}$ &89.00\\
\midrule
AVG & 80.46 & 82.01 & 81.76 & 82.72 & 82.24 & 82.67 & 83.45 &85.67 & $\mathbf{85.78}$\\
\bottomrule
\end{tabular}}
\caption{Multi-source unsupervised domain adaptation results on the Amazon review dataset.}
\label{tb-uda}
\end{table*}

%% file: tables/tb-para-comparision.tex
\begin{table}[]
\resizebox{0.5\textwidth}{!}{%
\begin{tabular}{@{}l|rr|rr@{}}
\toprule
Dataset     & \multicolumn{2}{c|}{Amazon}        & \multicolumn{2}{c}{FDU-MTL}          \\ \midrule
Model          & \multicolumn{1}{c|}{MAN}    & \multicolumn{1}{c|}{SAN(ours)} & \multicolumn{1}{c|}{MAN}     & \multicolumn{1}{c}{SAN(ours)} \\ \midrule
\# Para. of $F_s$ & \multicolumn{1}{r|}{5.57M} & 5.57M & \multicolumn{1}{r|}{20.20M} & 20.20M \\
\# Para. of $\{F_d^i\}_{i=1}^M$   & \multicolumn{1}{r|}{22.13M} & \textbf{5.57M}                 & \multicolumn{1}{r|}{322.65M} & \textbf{20.20M}               \\
\# Para. of $C$ & \multicolumn{1}{r|}{0.04M} & 0.04M & \multicolumn{1}{r|}{0.04M}  & 0.04M  \\
\# Para. of $D$ & \multicolumn{1}{r|}{0.02M} & 0.02M & \multicolumn{1}{r|}{0.02M}  & 0.02M  \\
\# Total Para. & \multicolumn{1}{r|}{27.76M} & \textbf{12.00M}                & \multicolumn{1}{r|}{342.91M} & \textbf{40.46M}               \\ \bottomrule
\end{tabular}%
}
\caption{Model parameter comparison between MAN and SAN (Ours)}
\label{tb-para-comparision}
\end{table}

%% file: tables/tb-runtime.tex
\begin{table}[]
\centering
\resizebox{0.28\textwidth}{!}{%
\begin{tabular}{@{}lcc@{}}
\toprule
Model     & Amazon         & FDU-MTL         \\ \midrule
MAN       & 7.07s          & 70.88s          \\
SAN(ours) & \textbf{6.39s} & \textbf{60.72s} \\ \bottomrule
\end{tabular}%
}
\caption{Model runtime comparison between MAN and SAN (ours)}
\label{tb-runtime}
\end{table}

%% file: Appendix/tables/tb-data-amazon.tex
\begin{table}[h]
\begin{center}
\resizebox{0.7\columnwidth}{!}{
\begin{tabular}{ l| c c c }
\toprule
Domain & Labeled & Unlabeled & Class.\\
\midrule
Books & 2000 & 4465 & 2\\
DVD & 2000 & 5681 & 2\\
Electronics & 2000 & 3586 & 2\\
Kitchen & 2000 & 5945 & 2\\
\bottomrule
\end{tabular}}
\end{center}
\caption{Details of the Amazon review dataset}
\label{tb-data-amazon}
\end{table}

%% file: Appendix/tables/tb-data-fdu.tex
\begin{table}[h]
\begin{center}
\resizebox{1\columnwidth}{!}{
\begin{tabular}{ l| c c c c c c c c c}
\toprule
Domain & Train & Dev. & Test & Unlabeled & Avg. L & Vocab. & Class.\\
\midrule
Books & 1400 & 200 & 400 & 2000 & 159 & 62K & 2\\
Electronics & 1398 & 200 & 400 & 2000 & 101 & 30K & 2\\
DVD & 1400 & 200 & 400 & 2000 & 173 & 69K & 2\\
Kitchen & 1400 & 200 & 400 & 2000 & 89 & 28K & 2\\
Apparel & 1400 & 200 & 400 & 2000 & 57 & 21K & 2\\
Camera & 1397 & 200 & 400 & 2000 & 130 & 26K & 2\\
Health & 1400 & 200 & 400 & 2000 & 81 & 26K & 2\\
Music & 1400 & 200 & 400 & 2000 & 136 & 60K & 2\\
Toys & 1400 & 200 & 400 & 2000 & 90 & 28K & 2\\
Video & 1400 & 200 & 400 & 2000 & 156 & 57K & 2\\
Baby & 1300 & 200 & 400 & 2000 & 104 & 26K & 2\\
Magazine & 1370 & 200 & 400 & 2000 & 117 & 30K & 2\\
Software & 1315 & 200 & 400 & 475 & 129 & 26K & 2\\
Sports & 1400 & 200 & 400 & 2000 & 94 & 30K & 2\\
IMDB & 1400 & 200 & 400 & 2000 & 269 & 44K & 2\\
MR & 1400 & 200 & 400 & 2000 & 21 & 12K & 2\\
\bottomrule
\end{tabular}}
\end{center}
\caption{Details of the FDU-MTL dataset}
\label{tb-data-fdu}
\end{table}

%% file: Appendix/tables/tb-verify.tex
\begin{table}[h]
\begin{center}
\resizebox{1\columnwidth}{!}{
\begin{tabular}{l| c c c c c }
  \toprule
  Method & Books & DVD & Elec. & Kit. & AVG \\
  \midrule
  pSAN & 82.25 & 83.05 &86.90 &88.25 &85.11 \\
  pSAN w/ $F_{d}^{zero}$ & 81.70 & 82.15 & 86.10 & 87.90 & 84.46 \\
  pSAN w/ $F_{d}^{shuf}$ & 81.10 & 82.00 & 85.90 & 87.35 & 84.09\\
  \bottomrule
\end{tabular}}
\end{center}
\caption{Validity verification of stochastic feature extractor on the Amazon review dataset}
\label{tb-verify}
\end{table}

%% file: Appendix/tables/tb-ablation-amazon.tex
\begin{table}[h]
\begin{center}
\resizebox{1\columnwidth}{!}{
\begin{tabular}{l| c c c c c }
\toprule
Domain & SAN(full) & SAN w/o dls & SAN w/o rplr & plain SAN \\
\midrule
Books   & 86.29 & 84.70 & 83.05 & 82.25 \\
DVD     & 86.43 & 85.10 & 83.35 & 83.05 \\
Electr. & 89.78 & 89.75 & 87.75 & 86.90 \\
Kit.    & 91.31 & 90.85 & 88.15 & 88.25 \\
AVG     & 88.45 & 87.60 & 85.53 & 85.11 \\
\bottomrule
\end{tabular}}
\end{center}
\caption{Ablation study analysis on the Amazon review dataset}
\label{tb-ablation-amazon}
\end{table}

%% file: Appendix/tables/tb-ablation-fdu.tex
\begin{table}[h]
\begin{center}
\resizebox{1\columnwidth}{!}{
\begin{tabular}{ l| c c c c  }
\toprule
Domain & SAN(full) & SAN w/o dls & SAN w/o rplr & plain SAN \\
\midrule
books & 90.5 & 89.0& 87.0 & 87.8 \\
electronics & 87.7 & 86.5 & 88.5 & 88.8 \\
dvd & 89.7 & 90.0 & 90.8 & 88.3 \\
kitchen & 90.4 & 90.3 & 90.5 & 89.8 \\
apparel & 87.4 & 86.0 & 87.5 & 87.3 \\
camera & 91.1 & 90.8 & 91.3 & 89.8\\
health & 90.3 & 90.5 & 90.0 & 91.3 \\
music & 85.9 & 86.5 & 85.3 & 85.8 \\
toys & 90.3 & 91.3 & 90.8 & 89.5 \\
video & 90.0 & 90.3 & 88.3 & 89.5 \\
baby & 90.7 & 90.8 & 90.0 & 90.0 \\
magazine & 92.3 & 91.8 & 93.0 & 92.3 \\ 
software & 89.5 & 89.0 & 90.5 & 89.0 \\
sports & 90.0 & 88.0 & 88.8 & 90.3 \\
IMDb & 89.3 & 89.8 & 88.8 & 86.5 \\
MR & 76.5 & 76.3 & 74.3 & 72.0  \\
\midrule
AVG & 88.8 & 88.5 & 88.4 & 88.0  \\
\bottomrule
\end{tabular} }
\end{center}
\caption{Ablation Study on the FDU-MTL dataset}
\label{tb-ablation-fdu}
\end{table}

%% file: Appendix/tables/tb-limit-comparison.tex
\begin{table}[h]
\begin{center}
\resizebox{1\columnwidth}{!}{
\begin{tabular}{ l| c c c c c c}
\toprule
Method & CAN & MRAN & CRAL & MBF & RCA & SAN(ours)\\
\midrule
Amazon & 87.70 & 87.64 & 88.00 & 87.71 & 86.88 & \textbf{88.45} \\
FDU-MTL & 89.4 & 89.0 & \textbf{90.2} & 90.1 & 89.0 & 88.8 \\
\bottomrule
\end{tabular}}
\end{center}
\caption{Comparisons of SAN with several state-of-the-art methods}
\label{tb-limit-comparision}
\end{table}

%% file: Appendix/tables/tb-limit-amzon.tex
\begin{table}[h]
\begin{center}
\resizebox{1\columnwidth}{!}{
\begin{tabular}{ l| c c c c c}
\toprule
Domain & Books  & DVD & Elec. & Kit. & AVG\\
\midrule
Acc. & 90.27 & 88.91 & 94.52 & 94.66 & $92.09\pm2.10$\\
\bottomrule
\end{tabular}}
\end{center}
\caption{Accuracy of valid pseudo-labels on the Amazon review dataset}
\label{tb-limit-amazon}
\end{table}

%% file: Appendix/tables/tb-limit-fdu.tex
\begin{table*}[h]
\begin{center}
\resizebox{2\columnwidth}{!}{
\begin{tabular}{ l| c c c c c c c c c c c c c c c c c}
\toprule
Domain & Books & Elec. & DVD  & Kit. & Apparel & Camera & Health & Music & Toys & Video & Baby & Magaz. & Softw. & Sports & IMDb & MR & AVG\\
\midrule
Acc. & 89.67 & 94.63 & 90.69 & 95.21 & 96.05 &94.84 &93.98 &87.76 &93.12 &90.25 &93.99 &95.82 &92.17 &95.88 &89.33 &82.87 &$92.27\pm3.52$\\
\bottomrule
\end{tabular}}
\end{center}
\caption{Accuracy of valid pseudo-labels on the FDU-MTL dataset}
\label{tb-limit-fdu}
\end{table*}